\documentclass[times,twocolumn,final,authoryear]{elsarticle}
\usepackage[utf8]{inputenc}
\usepackage{prletters}
\usepackage{framed,multirow}

\usepackage{graphicx}
\usepackage{color}
\usepackage{siunitx}

\usepackage{url}
\usepackage{xcolor}
\definecolor{newcolor}{rgb}{.8,.349,.1}
\journal{Pattern Recognition Letters}

\title{Convolutional Neural Network Simplification with Progressive Retraining}
\author[1]{Osaku, D.}
\ead{danosaku@hotmail.com}

\author[1]{Gomes, J.F.}
\ead{jgomes@ic.unicamp.br}

\author[1]{Falcão, A.X.}
\ead{afalcao@ic.unicamp.br}

 \address[1]{State University of Campinas, Brazil}

\date{March 2020}

\begin{document}

\begin{abstract}
    Kernel pruning methods have been proposed to speed up, simplify, and improve explanation of convolutional neural network (CNN) models.  However, the effectiveness of a simplified model is often below the original one. In this letter, we present new methods based on objective and subjective relevance criteria for kernel elimination in a layer-by-layer fashion. During the process, a CNN model is retrained only when the current layer is entirely simplified, by adjusting the weights from the next layer to the first one and preserving weights of subsequent layers not involved in the process. We call this strategy \emph{progressive retraining}, differently from kernel pruning methods that usually retrain the entire model after each simplification action -- e.g., the elimination of one or a few kernels. Our subjective relevance criterion exploits the ability of humans in  recognizing visual patterns and improves the designer's understanding of the simplification process. The combination of suitable relevance criteria and progressive retraining shows that our methods can increase effectiveness with considerable model simplification. We also demonstrate that our methods can provide better results than two popular ones and another one from the state-of-the-art using four challenging image datasets.
\end{abstract}


\maketitle
\section{Introduction}

The simplification of a pretrained neural network involves the elimination of neurons that might cause minor effectiveness loss after retraining (\cite{80236,lecun1990optimal}). These methods have been proposed for decades. When applied to the convolutional neural network (CNN) models,  they are often referred to as kernel pruning methods. Such methods can speed up, simplify, and improve the explanation of the models~(\cite{he2017channel, huang2018learning, augasta2013pruning}), but their effectiveness is often below the original one~(\cite{luo2017thinet,Jordao:2019, li2016pruning}). 

We have observed that kernel pruning methods usually retrain the entire model after each simplification action -- e.g., the elimination of one or a few kernels from a given layer. We call this strategy \emph{complete retraining}. It alters the weights of subsequent layers that have not been involved yet. Given the known problems of retraining deep neural networks, we presume that the accuracy loss in those simplified models is related to complete retraining and the criteria adopted for kernel elimination. 

\begin{figure}[!ht]
 \begin{center}
 \includegraphics[width=0.9\hsize]{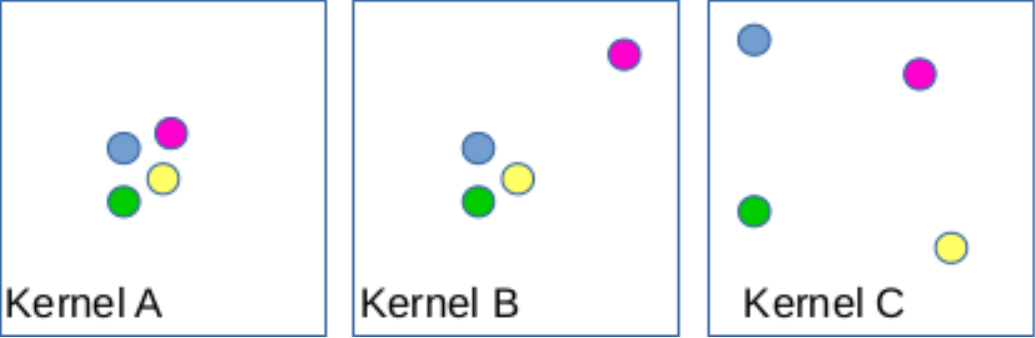}
 \end{center} 
 \caption{Hypothetical 2D projections~(\cite{maaten2008visualizing}) of the mean activation maps of three kernels (A, B, and C) for four color-coded classes. We remove kernel A and keep kernels B and C in the CNN model.}
 \label{f.subjective}
\end{figure}

In this letter, we circumvent the problem by presenting kernel pruning methods that operate in a layer-by-layer fashion with effectiveness improvements. The methods include new strategies for (a) kernel elimination and (b) model retraining after each layer simplification. In (a), we present \emph{objective} and \emph{subjective} kernel relevance criteria. In the objective criterion, lower is the negative impact of a kernel in the cross-entropy loss function higher is its priority for removal. In the subjective criterion, the user (a CNN designer) visualizes a 2D projection ~(\cite{maaten2008visualizing}) of the \emph{mean activation maps} that the kernels generate for each class, and eliminates kernels that cannot separate classes in the projection. Each kernel of a given convolutional layer produces one activation map for each input image. The mean activation map of images from the same class indicates the most relevant regions activated by that kernel. We expect that a relevant kernel activates differently for at least one class. Therefore, a kernel remains in the model when at least one class appears separated from the others in the projection of its mean activation maps (Figure~\ref{f.subjective}). This subjective criterion explores the ability of humans to recognize visual patterns and improves the expert's understanding of the model simplification process.  In (b), we present a new strategy, named \emph{progressive retraining}, in which the CNN model is retrained from the layer after the simplified one to layer one by backpropagation. Progressive retraining aims to recover the model's generalization capability before the simplification of the next convolutional layer. 

We first discuss related works and emphasize the importance of the above contributions (Section~\ref{sec:related_work}). Section~\ref{sec:methods} describes two new CNN simplification methods that operate in a layer-by-layer fashion based on the proposed kernel relevance criteria and progressive retraining. The experiments first evaluate the advantages of  progressive retraining over complete retraining on four challenging image datasets (Section~\ref{sec:exper}). These advantages come with significant efficiency gains over a reference CNN model, Vgg-16~(\cite{vggNet}), and improvements in effectiveness --- a rare phenomenon in kernel pruning methods. We finally show the effectiveness gains of the proposed approaches over two popular kernel pruning methods~(\cite{li2016pruning,hu2016network}) and one approach from the state-of-the-art~(\cite{Jordao:2019}).

\section{Related works}
\label{sec:related_work}

Neural network simplification methods have been investigated for decades.~\cite{80236} estimates the sensitivity of the global error function when removing neurons, and prunes low sensitivity connections.~\cite{549081} propose a similar idea by estimating sensitivity at the output of the layer after the one under simplification.~\cite{POPPI1998187} adopt a special algorithm to measure the impact of removing a neuron in the training error.~\cite{737502} use an extended Kalman filter to measure the importance of a connection weight in a network.~\cite{augasta2013pruning} provide a comparative study of neuron pruning methods on real datasets.

The above methods have been focused on multi-layer perceptron (MLP) models. More recently, after the introduction of CNN models, new kernel pruning methods have appeared. They aim at simplifying 
the CNN architecture by eliminating redundant kernels and/or kernels that can be removed with minor negative impact in accuracy. However, it has been difficult for those methods to obtain significant network simplifications with no accuracy loss
~(\cite{luo2017thinet,Jordao:2019, li2016pruning}). 

According to~\cite{he2017channel}, recent advances in CNN acceleration can be divided into three categories: (1) optimized implementations~(\cite{vasilache2014fast}), (2) weight quantization~(\cite{10.1007/978-3-319-46493-0_32}), and (3) structured simplification~(\cite{jaderberg2014speeding}). Kernel pruning methods fall into category (3).~\cite{lecun1990optimal}, for instance, identify irrelevant connection weights based on their negligeable impact in accuracy.~\cite{li2016pruning}  determine the importance of each kernel using the L1 norm of the weights and then kernels with lower values are pruned. By conducting simplification experiments on Vgg-16, ResNet34, ResNet56, and ResNet101, the authors report small losses in accuracy for reduction factors from $20\%$ to $30\%$ of the original number of kernels.~\cite{luo2017thinet} measure the impact of kernel removal at the output of the next layer to identify possible redundant kernels. Kernels that cause smaller changes in the total activation map of the next layer are selected for removal. They report a slight increase in the top-5 accuracy on the ImageNet dataset, when 50\% of the kernels in the first 10 layers of Vgg-16 are removed, and a slight decrease in the top-5 accuracy, when fully connected layers are replaced by general average pooling, which represents a considerable simplification of Vgg-16. Other kernel relevance criteria may be based on Taylor expansion~(\cite{DBLP:journals/corr/MolchanovTKAK16}) or the percentage of zero activations~(\cite{hu2016network}).

More complex simplification approaches have also been proposed. For instance,~\cite{huang2018learning} propose pruning agents to detect and eliminate unnecessary kernels. This pruning agent is modeled as a second CNN model which takes the kernel weights of the model under simplification as input and output binary decisions about removing or not kernels. The authors report a loss in accuracy of about 3\% after pruning the Vgg-16 and ResNet18 models using the CIFAR-10 dataset.~\cite{he2017channel} employ Lasso regression to determine redundant kernels for pruning and report slight accuracy losses with some simplified CNN models.

More recently,~\cite{Jordao:2019} estimates the importance of kernels for pruning based on the relationship between classes in a latent space, generated by partial least squares, and their variable importance in projection. They first eliminate kernels in the entire network and then retrain the model. They report considerable simplification without penalizing accuracy. 

We present kernel pruning methods that simplify the convolutional layers of a CNN model in a layer-by-layer fashion. Differently from all previous approaches, we retrain the model progressively, without affecting the weights of subsequent layers that have not been involved yet. We also introduce a subjective criterion based on the visual analysis of data projections. By that, we would like to call attention for the importance of using visual analytics in the design of deep neural networks. The strategy seems promising to guide human actions in the machine learning loop, such that the resulting models are as simple and effective as possible for a given problem. In this context, there is a lack of interactive methods for the construction of deep neural networks.  

Visual analytics have been successfully employed to understand neural networks and explain their decisions~(\cite{morch95,10.1007/978-3-540-24844-6_5,10.1007/978-3-319-10590-1_53,RauberTVCG2017,Hohman:2020}). It has also been used in the design of deep neural network models~(\cite{8019872,GarciaIJCNN19}).~\cite{8019872} present a system, named DeepEyes, to support the design of such models,  and~\cite{GarciaIJCNN19} show a first method to build a simplified CNN model in a layer-by-layer fashion. They introduce a kernel selectivity criterion based on the visual analysis of the \emph{activation occurrence maps} for each class. That is, by counting the positive activations for input images from each class, the user can visualize those activation maps as 2D images and choose the kernels with different activation patterns among classes. The method is limited by the number of classes and the 2D dimension of the activation maps (2D input images). In the proposed subjective criterion, we use 2D projections to allow more classes and overcome the limitation about the dimension of the activation maps. However, we are still limited to reasonable numbers of classes, layers, and kernels per layer.  

\section{Pruning with progressive retraining}
\label{sec:methods}

\begin{figure}[!ht]
 \begin{center}
 \includegraphics[width=\hsize]{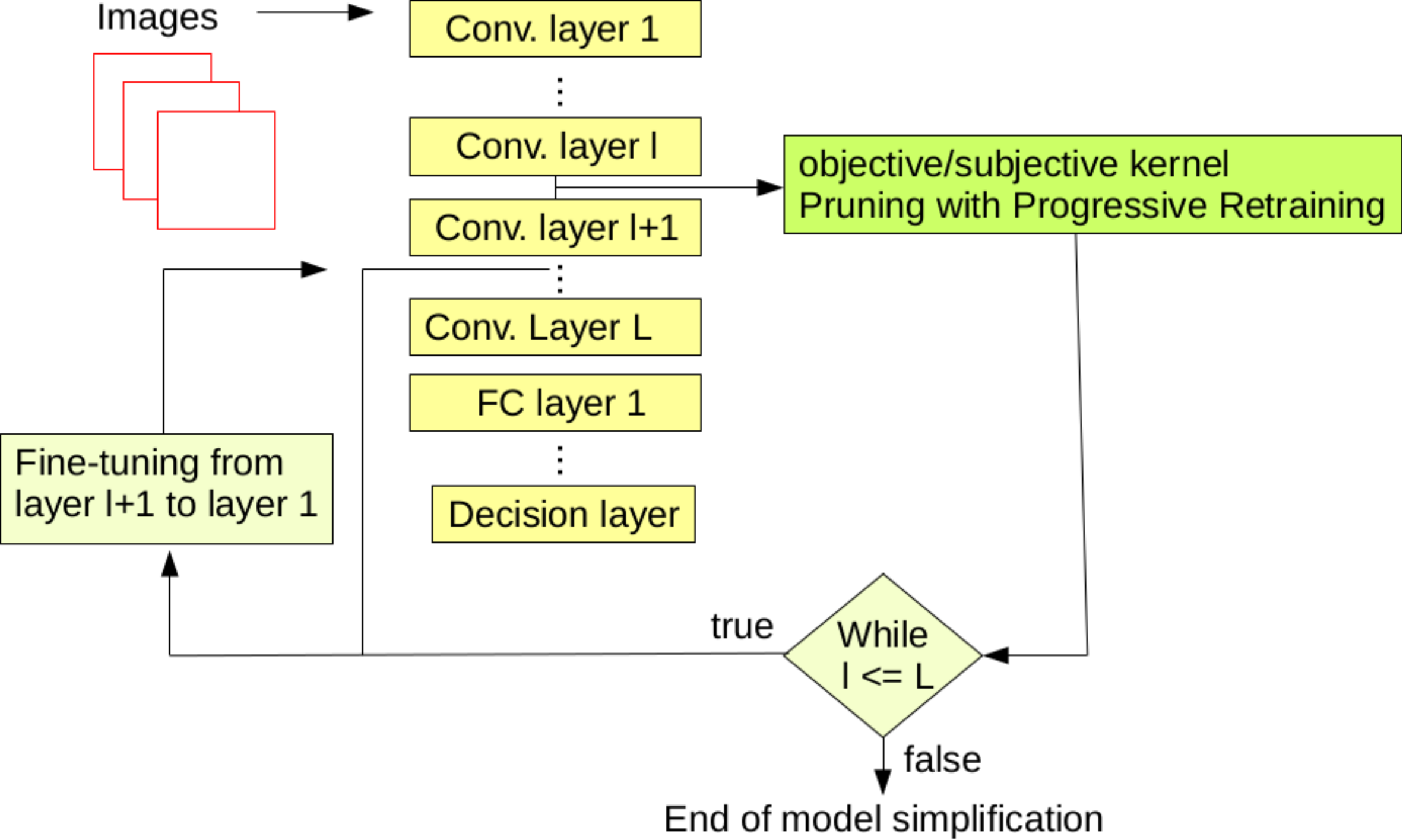}
 \end{center}
 \caption{Kernel pruning with progressive retraining using an objective/subjective criterion. Progressive retraining for a simplified layer $l$ starts from the output of layer $l+1$ to layer $1$ by backpropagation.}
 \label{f.PPR}
\end{figure}

This section presents two approaches to simplify CNNs based on kernel relevance criteria followed by progressive retraining (fine-tuning) in a layer by layer fashion (Figure~\ref{f.PPR}). One may conceptually divide a CNN for image classification into three parts: (1) a sequence of convolutional layers, with each layer containing convolution, activation, and optional operations (e.g., pooling, batch normalization) useful for feature extraction; (2) fully connected layers for feature space reduction, which discover the neurons specialized in each class~(\cite{RauberTVCG2017}); and (3) a decision layer for final classification, being (2) and (3) known as a MLP classifier. We have applied the following procedures to simplify pre-trained CNNs based on the analysis of each convolutional layer's output. 

\subsection{Objective pruning}

The first approach, named \emph{objective Pruning with Progressive Retraining} (oPPR), eliminates kernels of a given convolutional layer $l$ based on an \emph{objective relevance criterion} and fine-tunes the network by backpropagation during $x$ epochs (e.g., $x=40$) from layer $l+1$ to layer $1$, aiming  to recover its previous activations at the output of layer $l+1$. This process starts at layer $1$ and proceeds until the simplification of the last convolutional layer $L$, when the fine-tuning involves the weight optimization from the first fully connected layer $L+1$ to layer $1$. 

In this approach, the relevance of a kernel $k$ is related to its impact on the neural network's loss function. We use cross-entropy as loss function --- higher is the loss, when removing a kernel $k$ from a layer $l$, more critical the kernel $k$ is for layer $l$. Given that, for a layer $l$, one can remove either a given number of the most irrelevant kernels or the kernels whose relevance is below a given threshold. 

Once a set ${\cal M}_l\subset {\cal K}_l$ of kernels have been removed from the original kernel set ${\cal K}_l$ of layer $l$, we apply backprogation to refine the weights from layer $l+1$ to layer $1$. The loss function in this case is the mean square difference of the activation values obtained with and without ${\cal M}_l$ at the output of layer $l+1$. That is, let $I$ be an input image to the network, ${\cal X}$ be the set of training images, $A_{l+1}(I,{\cal K}_l)\in \Re^{n_{l+1}}$ be the output of layer $l+1$, after ReLu and, whenever the case, after pooling, with $n_{l+1}$ activation values, as a result from the input $I$, and $A_{l+1}(I,{\cal K}_l\backslash {\cal M}_l)\in \Re^{n_{l+1}}$ be the same without ${\cal M}_l$. The loss function for progressive retraining is defined as 
\begin{eqnarray}
\bar{D}_{l+1} & = & \frac{1}{|{\cal X}|} \sum_{\forall I \in {\cal X}} \|A_{l+1}(I,{\cal K}_l) - A_{l+1}(I,{\cal K}_l\backslash {\cal M}_l)\|_2. 
\end{eqnarray}

\subsection{Subjective pruning}

\begin{figure}[hbt!]
 \centering
 \includegraphics[scale=0.35]{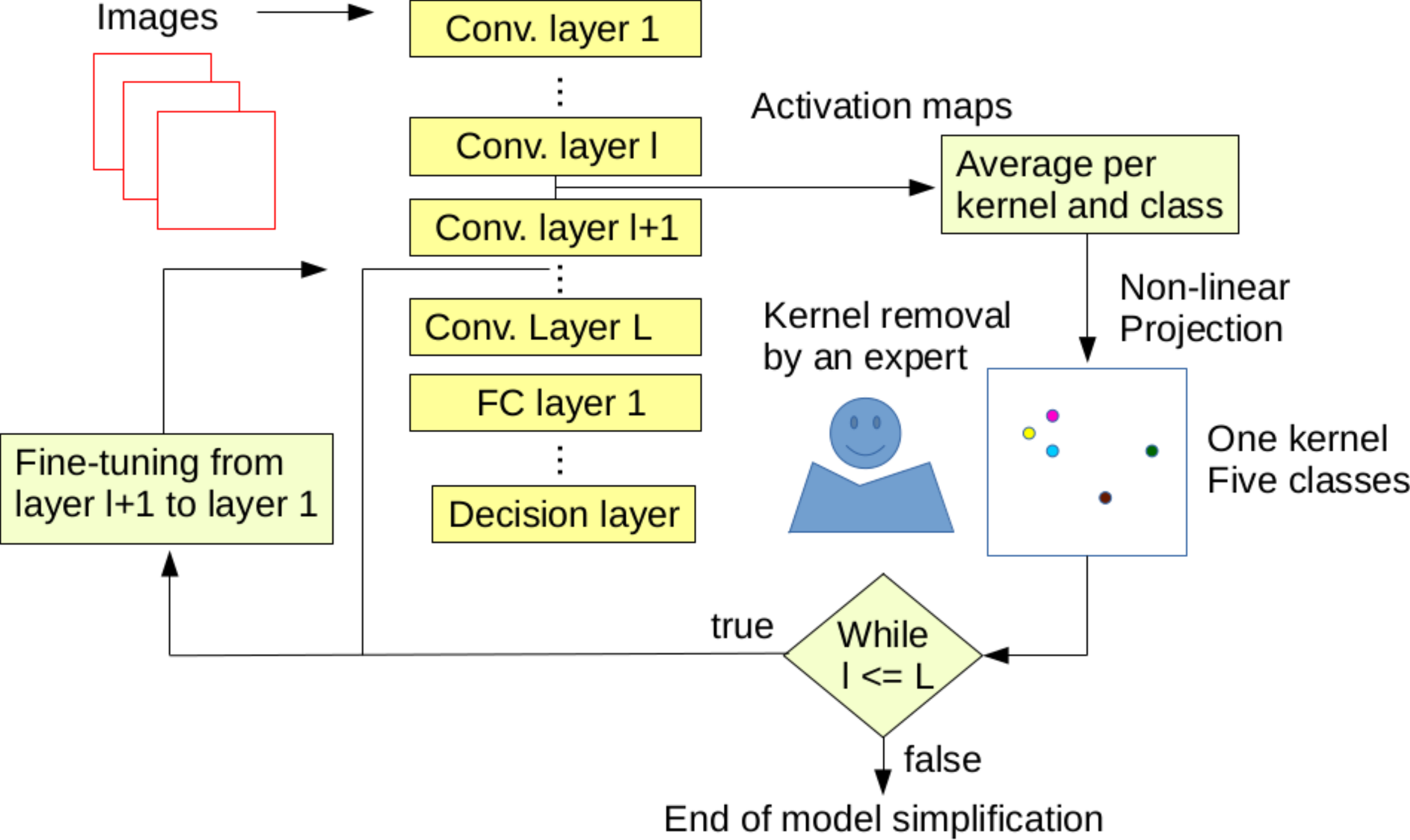}
 \caption{sPPR: for each convolutional layer $l$, the expert analyzes the projections of the mean activation maps of the classes from each kernel, and eliminates kernels that form one cluster of points in the projection.}
 \label{f.sPPR}
\end{figure}

The second approach, named \emph{subjective Pruning with Progressive Retraining} (sPPR), works similarly to oPPR, except for the kernel relevance criterion. In this case, an expert in the design of neural networks visualizes 2D non-linear projections of \emph{mean activation maps} and decides which kernels are the most relevant, according to the expert's subjective judgment of the separation in the projection among the mean activation maps of the classes for each given kernel. One should expect that a relevant kernel can separate at least one class from the others (Figure~\ref{f.sPPR}).  

Let ${\cal X}_j\subset {\cal X}$ be the set of training images from the class $j\in \{1,2,\dots,c\}$. Let $A_{l,k,j}(I)\in \Re^n$ be a map with $n < n_l$ activations at the output of layer $l$ as a consequence of using an input image $I$ from class $j$ and a kernel $k$ up to layer $l$. Note that these are the activations from the subset of the $n_l$ activations, which result from the convolution with kernel $k$ only. One mean activation map $\bar{A}_{l,k,j}$ from kernel $k$ and for each class $j$ can be obtained from the output of layer $l$ as
\begin{eqnarray}
\bar{A}_{l,k,j} & = & \frac{1}{|{\cal X}_j|}\sum_{\forall I\in {\cal X}_j}A_{l,k,j}(I). 
\end{eqnarray}

\begin{figure}[ht!]
 \begin{center}
\begin{tabular}{cc}     
 \fbox{\includegraphics[width=0.39\hsize]{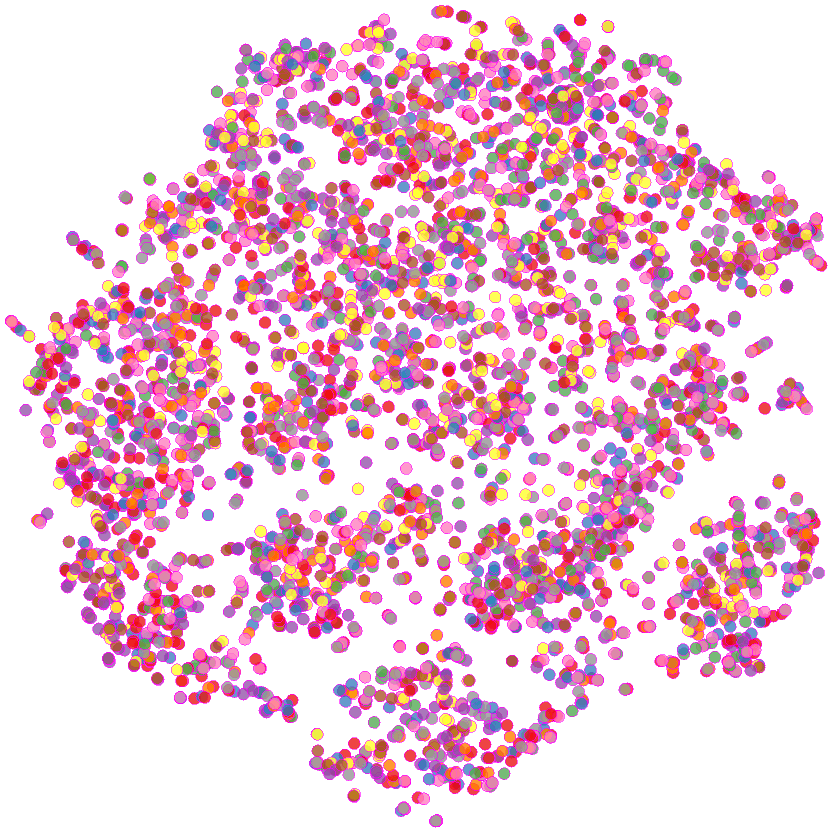}} & 
 \fbox{\includegraphics[width=0.39\hsize]{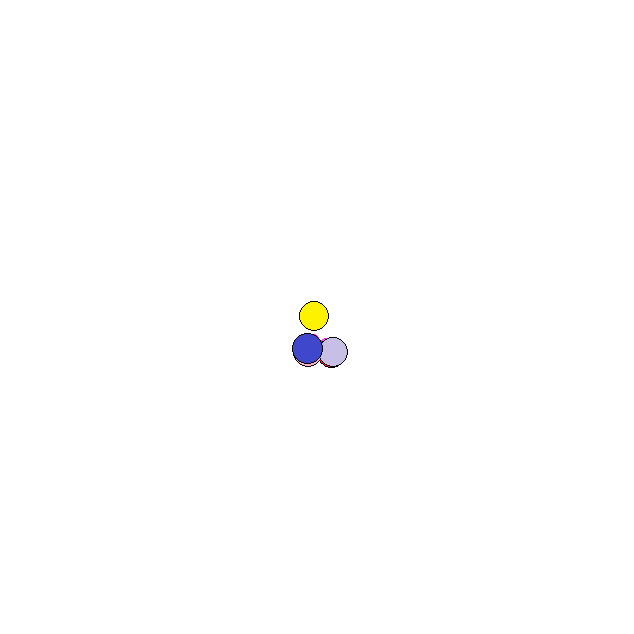}} \\
 (a) & (b) \\
 \fbox{\includegraphics[width=0.39\hsize]{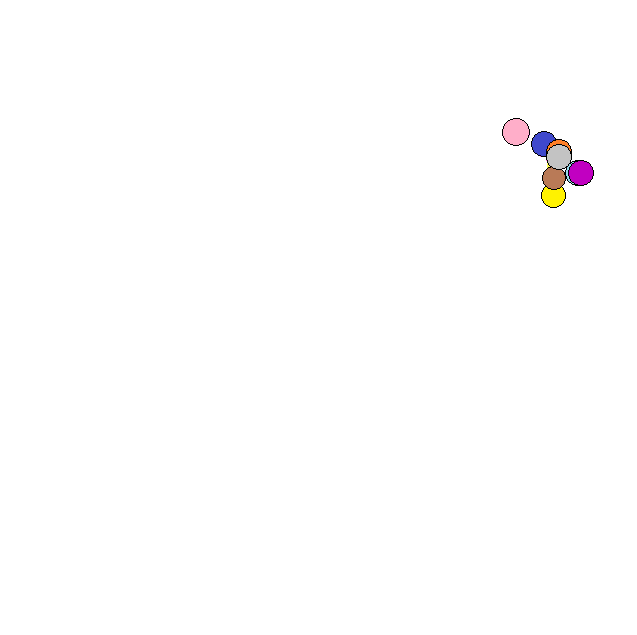}} &
 \fbox{\includegraphics[width=0.39\hsize]{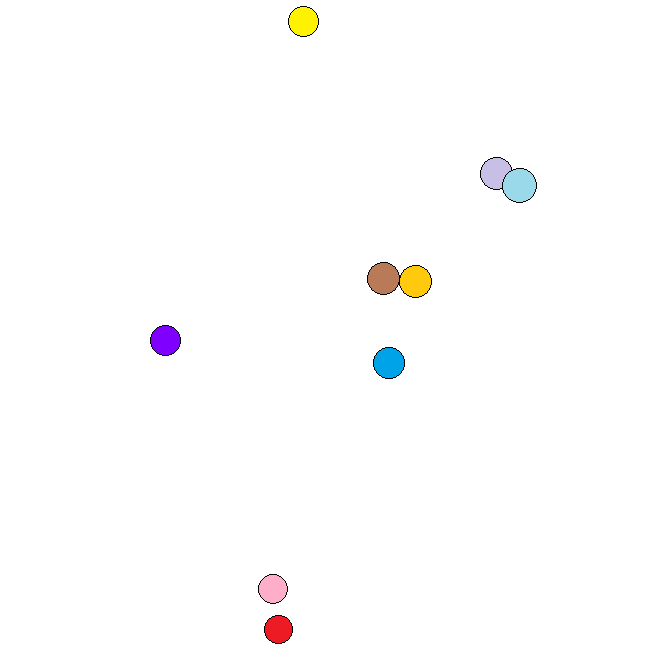}} \\
  (c) & (d)
 \end{tabular}
 \end{center} 
 \caption{For a layer with $512$ kernels and a dataset with $c=9$ classes. The t-SNE projection of the mean activation maps (a) from all kernels and (b-d) from three distinct kernels. The color indicates the class of images used to create the mean activation maps. The kernels in (b) and (c) are candidates for removal, while the kernel in (d) is a relevant one.}
 \label{f.kernelsel}
\end{figure}

We use the t-SNE algorithm~(\cite{maaten2008visualizing}) to project the mean activation maps $\bar{A}_{l,k,j}$ for all kernels $k$ and classes $j$ at the output of layer $l$ in the 2D space (Figure~\ref{f.kernelsel}a). Nevertheless, the expert analyzes the class projections of each kernel separately. For a reasonable number of kernels (e.g., 512), the user can quickly visualize only $c$ points in 2D per time from each kernel $k$ as an indication of how it discriminates among the images of the $c$ classes (Figures~\ref{f.kernelsel}b-\ref{f.kernelsel}d). If the points are very close to each other, forming one cluster in 2D, we may conclude that the kernel cannot discriminate between the classes. A kernel $k$ remains in layer $l$ when it can separate at least one class from the others. The separation between classes in the 2D projection is a subjective criterion that depends on the expert's visual analysis only.

After eliminating those irrelevant kernels from layer $l$, 
we apply backpropagation to refine the weights from layer $l+1$ to $1$, as described for the $oPPR$ approach. It is also possible to remove redundant kernels among the relevant ones. Redundant kernels have similar weights, so the projection of their weight vectors forms one cluster of points in 2D. Although this may further simplify the CNN, we have evaluated only the elimination of irrelevant kernels. 

\subsection{After network simplification}

After simplification, the entire simplified network is retrained with $y$ epochs (e.g., $y=50$). This process starts with random weights in the fully connected layers (MLP classifier)~(\cite{glorot2010understanding}) but only fine-tunes the weights of the convolutional layers, which resulted from progressive retraining. We have observed that the progressive retraining can recover the generalization capability of the CNN better than retraining the entire network after each layer simplification, as recommended in most works in~(\cite{li2016pruning,hu2016network,he2017channel,Jordao:2019}). 

\section{Experiments, results, and discussion}
\label{sec:exper}

To evaluate the simplification of CNNs using the $oPPR$ and $sPPR$ approaches, we divided the experiments into two parts. We use the Vgg-16~(\cite{vggNet}) model as reference, pre-trained on the ImageNet dataset~(\cite{imagenet_cvpr09}) and adjusted to the training set of each dataset below:
\begin{itemize}
    \item EGG-9: it is a private dataset with images of the eight most common species of helminth eggs in Brazil and one diverse class of fecal impurities.
    \item PRO-7: it is a private dataset with images of the six most common species of protozoa cysts in Brazil and one diverse class of fecal impurities.
    \item LAR-2: it is a private dataset with images of S. stecoralis (helminth larvae) and fecal impurities.
    \item STL-10: it is a well known public dataset with natural images, with considerable variations among objects that define a same class~(\cite{coates2011analysis}). 
\end{itemize}
In each dataset of intestinal parasites, EGG-9, PRO-7, and LAR-2, fecal impurities contain examples that are very similar in texture and shape to those of other classes, which makes these datasets very challenging. These datasets belong to a project for the automation of the diagnosis of intestinal parasites~(\cite{SuzukiTBME2013, Suzuki:2013:ISBI, SaitoPR2015, PeixinhoVipImage2015}).

For more statistically reliable results, we divided each dataset into five random splits of training and testing images, respectively, as described in Table~\ref{t.dataset}. To fine-tune Vgg-16 in each dataset, we used 50 epochs, learning rate $1e^{-5}$, and reinitialized the weights of its fully connected layers using the method in~(\cite{glorot2010understanding}).

\begin{table}
\footnotesize
\caption{Number of classes, pixels and bands (resolution), training samples, and testing samples in each dataset. \label{t.dataset}}
\begin{tabular}{|l|r|r|r|r|r|}
\hline
Dataset & Number of & Image      & Training & Testing & Total of \\ 
        & Classes   & Resolution & samples  & samples &  samples      \\ \hline
EGG-9   & 9         & 200x200x3  &   1265       &   10156   &  11421     \\ \hline
LAR-2   & 2         & 200x200x3  &   478        &   960     &    1438   \\ \hline
PRO-7   & 7         & 200x200x3  &    3733      &   29902   &   33635    \\ \hline
STL-10  & 10        & 96x96x3    &    3900      &    7800   &   11700    \\ \hline
\end{tabular}

\end{table}

We divide the experiments into two parts and discuss the pros and cons of oPPR and sPPR afterwards. 

\subsection{The impact of progressive retraining}

First, we evaluate the impact of the progressive retraining after each layer simplification by using variants of oPPR and sPPR, which adopt retraining of the entire network with a single epoch, as in other works~(\cite{li2016pruning,hu2016network,he2017channel,Jordao:2019}). For that, the variants of oPPR and sPPR with complete retraining (fine-tuning) after each layer simplification are named oPCR (\emph{Objective Pruning with Complete Retraining}) and sPCR (\emph{Subjective Pruning with Complete Retraining}), respectively.  Table~\ref{t.comb} presents the four approaches compared in this first experimental section.

\begin{table}
\centering
\caption{The proposed approaches, oPPR and sPPR, are first compared with their variants, oPCR and sPCR, that adopt retraining of the entire network after each layer simplification, as recomended in~\cite{he2017channel,Jordao:2019}.}
\label{t.comb}
\begin{tabular}{|l|c|c|}
\hline
Criterion  & \multicolumn{2}{c|}{Retraining}                               \\ \hline
          & \multicolumn{1}{l|}{Complete} & \multicolumn{1}{l|}{Progressive} \\ \hline
Objective  & oPCR                            & oPPR                               \\ \hline
Subjective & sPCR                           & sPPR                               \\ \hline
\end{tabular}

\end{table}

In oPPR and sPPR, we use $40$ epochs for progressive retraining and fixed learning rates along the epochs. To fine-tune from layer $l+1$ to layer $1$, the learning rate is fixed as described in Table~\ref{t.lr}, depending on the number of kernels of Vgg-16 at each convolutional layer $l$. Lower learning rates are used as higher is the number of kernels. 

After network simplification, the fully convolutional layers (the MLP classifier) are reinitialized with random weights~(\cite{glorot2010understanding}) for retraining and the weights of the simplified convolutional layers are refined. This retraining process uses 50 epochs and learning rate $1e^{-5}$ in all approaches: oPPR, sPPR, oPCR, and sPCR.

\begin{table}
\centering
\caption{Learning rates for progressive retraining.}
\label{t.lr}
\begin{tabular}{cccc}
\hline
\multicolumn{1}{|r|}{Layer} & \multicolumn{1}{l|}{Name}            & \multicolumn{1}{r|}{N. of Kernels} & \multicolumn{1}{r|}{Learn. rate}    \\ \hline
\multicolumn{1}{|l|}{1}     & \multicolumn{1}{l|}{block1\_conv\_1} & \multicolumn{1}{l|}{64}           & \multicolumn{1}{l|}{$1e^{-5}$} \\ \hline
\multicolumn{1}{|l|}{2}     & \multicolumn{1}{l|}{block1\_conv\_2} & \multicolumn{1}{l|}{64}           & \multicolumn{1}{l|}{$1e^{-6}$} \\ \hline
\multicolumn{1}{|l|}{3}     & \multicolumn{1}{l|}{block2\_conv1}   & \multicolumn{1}{l|}{128}          & \multicolumn{1}{l|}{$1e^{-6}$} \\ \hline
\multicolumn{1}{|l|}{4}     & \multicolumn{1}{l|}{block2\_conv2}   & \multicolumn{1}{l|}{128}          & \multicolumn{1}{l|}{$1e^{-7}$} \\ \hline
\multicolumn{1}{|l|}{5}     & \multicolumn{1}{l|}{block3\_conv1}   & \multicolumn{1}{l|}{256}          & \multicolumn{1}{l|}{$1^{e-7}$} \\ \hline
\multicolumn{1}{|l|}{6}     & \multicolumn{1}{l|}{block3\_conv2}   & \multicolumn{1}{l|}{256}          & \multicolumn{1}{l|}{$1^{e-7}$} \\ \hline
\multicolumn{1}{|l|}{7}     & \multicolumn{1}{l|}{block3\_conv3}   & \multicolumn{1}{l|}{256}          & \multicolumn{1}{l|}{$1e^{-8}$} \\ \hline
\multicolumn{1}{|l|}{8}     & \multicolumn{1}{l|}{block4\_conv1}   & \multicolumn{1}{l|}{512}          & \multicolumn{1}{l|}{$1e^{-8}$} \\ \hline
\multicolumn{1}{|l|}{9}     & \multicolumn{1}{l|}{block4\_conv2}   & \multicolumn{1}{l|}{512}          & \multicolumn{1}{l|}{$1e^{-8}$} \\ \hline
\multicolumn{1}{|l|}{10}    & \multicolumn{1}{l|}{block4\_conv3}   & \multicolumn{1}{l|}{512}          & \multicolumn{1}{l|}{$1e^{-8}$} \\ \hline
\multicolumn{1}{|l|}{11}    & \multicolumn{1}{l|}{block5\_conv1}   & \multicolumn{1}{l|}{512}          & \multicolumn{1}{l|}{$1e^{-8}$} \\ \hline
\multicolumn{1}{|l|}{12}    & \multicolumn{1}{l|}{block5\_conv2}   & \multicolumn{1}{l|}{512}          & \multicolumn{1}{l|}{$1e^{-8}$} \\ \hline
\multicolumn{1}{|l|}{13}    & \multicolumn{1}{l|}{block5\_conv3}   & \multicolumn{1}{l|}{512}          & \multicolumn{1}{l|}{$1e^{-8}$} \\ \hline
\end{tabular}

\end{table}

Tables~\ref{t.eggc}-\ref{t.stlc} present the results of comparison among the four approaches on the datasets EGG-9, PRO-7, LAR-2, and STL-10, respectively. One may observe that, except for two cases (oPPR in EGG-9 and sPPR in LAR-2), oPPR and sPPR could simplify Vgg-16 with some effectiveness gain (in accuracy and Cohen kappa) and considerable efficiency gain, as measured by the percentage of reduction in the number of kernels and GFLOPS. For the objective approaches, progressive retraining (oPPR) is consistently better than complete retraining (oPCR). In the subjective approaches, the expert is usually more restrained and eliminates considerably fewer kernels than the objective approaches. Nevertheless, the reduction in the number of kernels and GFLOPS with effectiveness gain is still significant in sPCR and sPPR.

\begin{table}
\scriptsize
\centering
\caption{Impact of the progressive retraining (oPPR and sPPR) after each layer simplification over complete retraining (oPCR and sPCR) and the original network (Vgg-16) on the EGG-9 dataset.}
\label{t.eggc}
\begin{tabular}{|l|l|l|l|l|}

\hline
     & Accuracy     & Cohen Kappa    & GFLOPs & Kernel   \\ 
     &      &     & reduction (\%) & reduction (\%)  \\ \hline  
Vgg-16 & 97.65  $\pm$  0.31 &  94.00  $\pm$  0.79   &   0.00  &  0.00  \\\hline
oPCR          & 96.90  $\pm$  0.24 &  92.02  $\pm$  1.09 &     74.42   & 50.00    \\\hline
oPPR &    97.62 $\pm$  0.10 &  93.94   $\pm$  0.26   &     74.42 & 50.00  \\\hline
sPCR  &     97.83 $\pm$ 0.01  &  94.43 $\pm$ 0.05    &   27.33  & 9.42 \\\hline
sPPR &    97.79 $\pm$ 0.20       &  94.34 $\pm$ 0.53 &  54.53 &  21.44\\\hline        
\end{tabular}

\end{table}

\begin{table}
\scriptsize
\centering
\caption{Impact of the progressive retraining (oPPR and sPPR) after each layer simplification over complete retraining (oPCR and sPCR) and the original network (Vgg-16) on the PRO-7 dataset.}
\label{t.proc}
\begin{tabular}{|l|l|l|l|l|}
\hline
    & Accuracy      & Cohen Kappa    & GFLOPs & Kernel   \\ 
     &      &     & reduction (\%) & reduction (\%)  \\ \hline  
Vgg-16  &   96.74  $\pm$ 0.15  &  91.93 $\pm$  0.36      &   0.00  & 0.00 \\\hline
oPCR         &   95.93  $\pm$ 0.48  &  89.91 $\pm$  1.15      &   74.42    & 50.00 \\\hline
oPPR  &   97.11 $\pm$  0.18  &    92.83 $\pm$ 0.41      &  74.42           & 50.00\\\hline
sPCR   &  97.01 $\pm$ 0.37  &    92.61 $\pm$ 0.88        &  29.37        & 10.52 \\\hline
sPPR   &    97.05 $\pm$ 0.21       &   92.68 $\pm$ 0.54  & 36.64          & 12.06\\\hline        
\end{tabular}

\end{table}

\begin{table}
\scriptsize
\centering
\caption{Impact of the progressive retraining (oPPR and sPPR) after each layer simplification over complete retraining (oPCR and sPCR) and the original network (Vgg-16) on the LAR-2 dataset.}
\label{t.larc}
\begin{tabular}{|l|l|l|l|l|}
\hline
  
    & Accuracy      & Cohen Kappa    & GFLOPs & Kernel   \\ 
     &      &     & reduction (\%) & reduction (\%)  \\ \hline  
Vgg-16  &   97.33  $\pm$ 0.81  &  89.65 $\pm$  3.29      &   0.00  &  0.00\\\hline
oPCR          &   96.24  $\pm$ 0.89  &  85.83 $\pm$  2.70      &   74.42  &  50.00\\\hline
oPPR  &   97.49 $\pm$  0.50 &    90.38 $\pm$ 2.03      &  74.42  & 50.00\\\hline
sPCR   &  97.39 $\pm$ 0.27  &    90.01 $\pm$ 1.08       &  38.07  & 16.41 \\\hline
sPPR   &    97.22 $\pm$ 0.66       &   89.32 $\pm$ 2.87  & 63.93 & 27.82 \\\hline        
\end{tabular}

\end{table}

\begin{table}
\scriptsize
\centering
\caption{Impact of the progressive retraining (oPPR and sPPR) after each layer simplification over complete retraining (oPCR and sPCR) and the original network (Vgg-16) on the STL-10 dataset.}
\label{t.stlc}
\begin{tabular}{|l|l|l|l|l|}
\hline
    & Accuracy      & Cohen Kappa    & GFLOPs & Kernel   \\ 
     &      &     & reduction (\%) & reduction (\%)  \\ \hline  
Vgg-16  &   77.89  $\pm$ 2.13  &  75.43 $\pm$  2.37      &   0.00  &  0.00\\\hline
oPCR          &   73.68  $\pm$ 0.98 &  70.75 $\pm$  1.09      &   74.25  &  50.00\\\hline
oPPR  &    79.24 $\pm$  1.74 &    76.93 $\pm$ 1.94     &  74.25 & 50.00 \\\hline
sPCR   &  80.95 $\pm$ 2.15  &    78.83 $\pm$ 2.39       &  0.20 & 3.85 \\\hline
sPPR   &    79.81 $\pm$ 1.33       &   77.57 $\pm$ 1.48  & 12.61 & 5.97 \\\hline        
\end{tabular}

\end{table}

\subsection{Comparison with other methods}

In this section, we compare the proposed approaches (oPPR and sPPR) with other network simplification techniques. These techniques usually differ in two important aspects: (a) the relevance criterion to eliminate kernels and (b) the retraining strategy of the network after each layer simplification. We selected the methods in~\cite{li2016pruning},~\cite{hu2016network},  and~\cite{Jordao:2019} for our comparative analysis. In~\cite{li2016pruning}, the relevance of a kernel is measured by the sum of the absolute values of its weights. In~\cite{hu2016network}, the kernels that generate more outputs with no activation are considered less relevants. In~\cite{Jordao:2019}, the kernels are selected using partial least squares and variable importance in projection. For a fair comparison, the kernels were pruned using iterative pruning with 5 iterations of $10\%$ of pruning ratio, which reduces in $34.39\%$ the number of kernels in the convolutional layers. All baselines adopted their original protocols for kernel pruning followed by complete retraining. The first two network simplification approaches are popular baselines for several works. 

In our case, we have fixed at $50\%$ the percentage of reduction in the number of kernels per layer for the objective approach, and repeated the results of the previous experiment for sPPR and the original Vgg-16, as references. For each layer, we eliminate 50\% of irrelevant kernels and apply progressive retraining. 

At the end of the simplification process, all simplified models are retrained with $50$ epochs, fixed learning rate $1e^{-5}$, and reinitialized fully connected layers~(\cite{glorot2010understanding}). 

Tables~\ref{t.acc50} and ~\ref{t.kappa50} show the results of this experiment on each dataset using accuracy and Cohen kappa as effectiveness measures, respectively. The results indicate that oPPR (and sPPR with less efficiency gain than oPPR) can be considerably more effective than the baselines, especially in Cohen kappa. This result is particularly important, because these datasets are unbalanced and Cohen kappa can penalize errors in small classes.

\begin{table}
\scriptsize

\caption{Mean accuracy with the elimination of $50\%$ of the kernels in the convolutional layers of Vgg-16. The best results are shown in bold.}
\label{t.acc50}

\begin{tabular}{|c|l|l|l|l|}
\hline
     & EGG-9 & PRO-7 & LAR-2 & STL-10 \\ \hline
Vgg-16  & 97.65  $\pm$  0.31    & 96.74  $\pm$ 0.15     & 97.33  $\pm$  0.81   &  77.89 $\pm$   2.13   \\ \hline
  
~\cite{li2016pruning}    & 94.48  $\pm$  0.66    & 93.29  $\pm$  0.46    &  95.98  $\pm$ 0.66   &  74.54  $\pm$ 2.19     \\ \hline

~\cite{hu2016network}    & 96.90  $\pm$  0.21    &  96.03 $\pm$  0.34   & 95.6  $\pm$  0.81   &  73.55  $\pm$  1.08    \\ \hline

~\cite{Jordao:2019}    &  96.71 $\pm$  0.47  & 94.83  $\pm$  0.46  &  96.85 $\pm$  0.30   &    75.89 $\pm$  1.18  \\ \hline

oPPR  & \textbf{97.62 $\pm$  0.10}    & \textbf{97.11 $\pm$  0.18}   & \textbf{97.49 $\pm$  0.50}   &  \textbf{79.24 $\pm$  1.74}   \\ \hline
sPPR   & \textbf{97.79 $\pm$ 0.20}    & \textbf{97.05 $\pm$ 0.21}   & \textbf{97.22 $\pm$ 0.66 }   &  \textbf{79.81 $\pm$ 1.33}   \\ \hline
\end{tabular}-

\end{table}

\begin{table}
\scriptsize
\caption{Mean Cohen kappa with the elimination of $50\%$ of the kernels in the convolutional layers of Vgg-16.}
\label{t.kappa50}
\begin{tabular}{|c|l|l|l|l|}
\hline
       & EGG-9 & PRO-7 & LAR-2 & STL-10 \\ \hline
 Vgg-16  &  94.00  $\pm$  0.78   &  91.93 $\pm$  0.36    & 89.65  $\pm$  3.29   & 75.43   $\pm$  2.37   \\ \hline

~\cite{li2016pruning}     & 85.51   $\pm$  1.89   &  83.39 $\pm$  0.83    &  84.30  $\pm$ 2.47   &  71.72  $\pm$ 2.44     \\ \hline
            
~\cite{hu2016network}   & 92.01   $\pm$  0.61    &  90.19 $\pm$  0.80    & 83.26  $\pm$  3.46   &   70.61 $\pm$ 1.20   \\ \hline

~\cite{Jordao:2019}     &  91.61  $\pm$  1.24  & 87.40  $\pm$   0.98   &  87.84 $\pm$  1.35   &   73.21 $\pm$  1.32 \\ \hline

oPPR   & \textbf{93.94   $\pm$  0.26}   &  \textbf{92.83 $\pm$ 0.41} &  \textbf{90.38 $\pm$ 2.03}   &  \textbf{76.93 $\pm$ 1.94}    \\ \hline
sPPR & \textbf{94.34 $\pm$ 0.53}   &   \textbf{92.68 $\pm$ 0.54} &  \textbf{89.32 $\pm$ 2.87}   &   \textbf{77.57 $\pm$ 1.48}     \\ \hline
\end{tabular}

\end{table}

\subsection{Discussion}
The experiments have demonstrated that the proposed progressive retraining can provide higher effectiveness gains than the complete retraining usually adopted in network simplification~(\cite{li2016pruning,hu2016network,he2017channel,Jordao:2019}). When simplifying networks, it is typical for the simplified version to be less effective than the original one, as observed for the methods in~(\cite{li2016pruning,hu2016network,Jordao:2019}) in all datasets. On the other hand, oPPR and sPPR,  consistently presented effectiveness gains over Vgg-16 (being oPPR in EGG-9 and sPPR in LAR-2, the exceptions for 50\% of kernel elimination). Given the considerable reduction in the number of kernels and GFLOPS, we may conclude that oPPR and sPPR are relevant contributions to the literature of network simplification. 

Although oPPR,  with a more aggressive kernel elimination, sometimes presented better performance than sPPR, the expert involvement in deep learning has several advantages. First, the expert can better understand the role of each element of the model and, perhaps, better explain its decisions~(\cite{RauberTVCG2017}). Second, the expert can intervene in the project of the network, as demonstrated by sPPR. Visual analytics techniques, such as t-SNE, play a central role in facilitating communication between machines and humans. They are known as essential tools to improve understanding of the machine learning process. However, we believe they can go beyond and let the user intervene to improve the machine learning process~(\cite{RauberInfoVis2018,BenatoSIBGRAPI2018}). In this sense, sPPR can improve with additional information about objective relevance measures to guide the expert's actions when eliminating kernels. A drawback in sPPR, however, is the limitation to reasonable numbers of classes, layers, and kernels per layer. On the other hand, oPPR completely relies on the objective relevance measure and eliminates a given and fixed percentage of kernels per layer. It seems that, with the expert in the process, some better criterion could be devised to eliminate an adaptive number of irrelevant kernels per layer.            

\section{Conclusion}

We have presented two solutions for CNN simplification, named oPPR and sPPR, which explore objective and subjective kernel relevance criteria, respectively, and perform progressive retraining to adjust the model's weights by preserving the weights of subsequent layers not involved in the simplification process. Progressive retraining has shown improvements over complete retraining, usually adopted in kernel pruning methods. The proposed methods have achieved considerable network simplification with effectiveness gains over the original model. They can also be more effective than two popular methods and one approach from the state-of-the-art. We may then conclude that oPPR and sPPR are relevant contributions to the literature of kernel pruning methods. 

We are now interested in investigating methods to construct CNNs in a layer-by-layer fashion by exploring objective and subjective kernel relevance criteria. In the context of the diagnosis of intestinal parasites, we intend to create more effective models with considerably simplified architectures.

\section{Acknowledgments}
 
The authors thank financial support from FAPESP (Proc. 2017/12974-0 and 2014/12236-1) and CNPq (Proc. 303808/2018-7).

\bibliographystyle{model2-names}
\bibliography{references}

\end{document}